\documentclass[conference]{IEEEtran}

\makeatletter
\newcommand{\linebreakand}{%
  \end{@IEEEauthorhalign}
  \hfill\mbox{}\par
  \mbox{}\hfill\begin{@IEEEauthorhalign}
}
\makeatother

\usepackage[utf8]{inputenc} 
\usepackage[T1]{fontenc} 
\usepackage{hyperref} 
\usepackage{url} 
\usepackage{booktabs} 
\usepackage{amsfonts} 
\usepackage{nicefrac} 
\usepackage{microtype} 
\usepackage{lipsum}
\usepackage{todonotes}
\usepackage{subcaption}

\title{A gamified simulator and physical platform for self-driving algorithm training and validation}

\begin{document}

\author{\IEEEauthorblockN{Joshua E. Siegel\IEEEauthorrefmark{1}}
\IEEEauthorblockA{Department of Computer Science and Engineering\\
Michigan State University\\
East Lansing, MI 48824\\
Corresponding author: jsiegel@msu.edu}
\and
\IEEEauthorblockN{Georgios Pappas\IEEEauthorrefmark{1}}
\IEEEauthorblockN{Konstantinos Politopoulos}
\IEEEauthorblockA{Department of Electrical and Computer Engineering \\
 National Technical University of Athens\\
 Athens, Greece}
\linebreakand
\IEEEauthorblockN{Yongbin Sun}
\IEEEauthorblockA{Department of Mechanical Engineering\\
Massachusetts Institute of Technology\\
Cambridge, MA 02139}
\\ \IEEEauthorrefmark{1} These authors contributed equally.
}

\maketitle

\IEEEpeerreviewmaketitle

\begin{abstract}
We identify the need for a gamified self-driving simulator where game mechanics encourage high-quality data capture, and design and apply such a simulator to collecting lane-following training data. The resulting synthetic data enables a Convolutional Neural Network (CNN) to drive an in-game vehicle. We simultaneously develop a physical test platform based on a radio-controlled vehicle and the Robotic Operating System (ROS) and successfully transfer the simulation-trained model to the physical domain without modification. The cross-platform simulator facilitates unsupervised crowdsourcing, helping to collect diverse data emulating complex, dynamic environment data, infrequent events, and edge cases. The physical platform provides a low-cost solution for validating simulation-trained models or enabling rapid transfer learning, thereby improving the safety and resilience of self-driving algorithms. 
\end{abstract}


\section{Introduction}
Deep Learning requires Big Data to identify subtle patterns, edge cases, or anomalies through increased exposure. One application making use of Big Data is vehicle automation. 

Self-driving is a growing field, and varied information is necessary to train resilient algorithms. Though the technical foundation for automation exists, engineers lack training data, particularly for infrequent events. 

Some manufacturers have access to an incumbent fleet's data (e.g. Tesla capturing image and telemetry data from customer vehicles\cite{tesla_waymo_1,tesla_waymo_2}). It's difficult to validate that these data are ``clean?? (for example, a drunk driver's road images might be treated as valid and negatively impact a lane-holding algorithm).``Wisdom of the crowd'' requires \textit{massive} scale if it is to be used in safety-critical systems. Other companies hire trained drivers, but professional drivers are expensive to hire. 

A lack of data limits self-driving research. There is a need for low-cost, high-quality data collection, and for inexpensive physical test platforms. This manuscript proposes gamified simulation as a means of collecting bulk data for self-driving and commodity hardware for algorithm validation. Specifically, we create a driving game where players compete for high scores or the best time to collect training data for line-following. The game's scoring mechanism provides an objective function encouraging users to collect ``good?? data. The result is an inexpensive human-in-the-loop simulation enabling rapid data collection. We also develop a low-cost physical test platform and prove the simulator's utility by training a model on synthetic data and effectively transferring that model to the physical domain.

The uniqueness of our solution is that the game enforces overt and latent rules in data collection. We learn from humans who have adapted to drive well in complex scenarios, while ensuring the data collected are of high quality. Altering game scoring mechanics can encourage users to collect information for both common and edge-case scenarios, yielding valuable training insight for low-frequency, high-risk events. The physical platform democratizes access to self-driving data collection and allows for rapid model validation for budget-constrained developers. 

\section{Prior Art}
\subsection{Simulation and Simluation}
Simulation and games are valuable data sources for deep learning\cite{Shafaei2016} with virtual worlds having been used to successfully train AI\cite{Johnson-Roberson2016}. However, computer-controlled simulations may not generate sufficiently-diverse data for models to observe rare events.

In self-driving, Google's Waymo uses simulation to artificially increase training data diversity\cite{tesla_waymo_1}. However, generated data abide by implicit and explicit simulation rules, meaning algorithms may learn latent rules that do not accurately mirror reality and may miss ``long tail'' events\cite{tesla_long_tail}. Though Waymo's vehicles have reduced crash rates relative to those of human drivers\cite{Teoh2017}, there is room for improvement. Real-world scenarios suffer from the entropy and chaos inherent in physical systems operated by irrational agents.

To capture unpredictable events, Waymo, Tesla and others collect real-world data from highly-instrumented fleets\cite{tesla_waymo_1,tesla_waymo_2}, but struggle to label samples. This complexity, and the cost of streaming large-volume data, make capturing complex scenarios unlikely -- and it is precisely these low-frequency, high-impact situations where automated vehicles struggle to match human performance (models are good at \textit{remembering} data, not \textit{reasoning}). 

In software, a common aphorism is the ``ninety-ninety'' rule, which states that ``the first $90$\% of the code accounts for the first $90$\% of the development time. The remaining $10$\% of the code accounts for the other $90$\% of the development time.'' This holds true for self-driving data capture -- the most complex and rarest inputs are left until the end to capture, but these are among the most critical to the system's performance. Advanced simulations are critical to addressing the ``last $10$\%'' of data collection. 

To this end, researchers have used video games to increase the volume of labeled training data for self-driving algorithms. The use of games lowers the cost of data capture relative to physical driving, with parallel play making it common for networks to observe and learn from infrequent events. 

Researchers modified Grand Theft Auto V (GTA V)\cite{Games} to capture information including vehicle speed, steering angle, and synthetic camera data \cite{Martinez2017}. Franke (2017) was able to use data to train radio controlled vehicles to drive\cite{Franke2017}. However, GTA V lacks output flexibility, e.g. with respect to matching vehicle or sensor parameters to physical systems, and samples require manual supervision. 

Fridman's DeepTraffic (2018) is designed with algorithm development in mind\cite{Fridman2018}. However, DeepTraffic generates information from a simplified 2D world. While DeepTraffic crowdsources data, it does not crowdsource \textit{human-operated} training data. 

The open source driving simulator VDrift \cite{VDrift} was used to create an optical flow dataset used to train a pairwise CRF model for image segmentation\cite{Haltakov2013}. However, the tool used for simulation was not designed with reconfigurability or flexible data output in mind (aside from being open source). 

Another gamified simulation is SdSandbox and its derivatives\cite{Kramer,Yu,Simulator}, which generate steering/image pairs from virtual vehicles and environments. These tools are flexible but generate unrealistic images, may or may not be human-in-the loop, and are not designed to crowdsource training data from multiple human drivers -- a feature which dramatically improves data capture rate and diversity. 

CARLA\cite{Dosovitskiy17} is a game-engine based cross-platform simulator emulating self-driving sensors and offering programmable traffic and pedestrian scenarios. While CARLA is a useful and flexible research tool, it lacks intrinsic motivation for users to collect clean data, limiting the ability to trust data ``cleanliness.'' Further, there is no physical analog to the in-game vehicle. A solution with game mechanics motivating user performance and a low-cost physical test platform would add significant research value. 

Other, task-specific elements of automated driving have been demonstrated using both simulators and games, e.g. pedestrian detection \cite{Marin2010,Hattori2015} and stop sign detection\cite{Filipowicz2017LearningTR}. Some simulators enable evolutionary computing, testing transferrability of driving skills across varied virtual environments. \cite{1688444} These approaches have not been integrated with physical testbeds, which could yield insight into real-world operations. These tools prove that simulated data can be used to effectively train self-driving algorithms.

There is an opportunity to create a gamified simulator and physical platform for collecting self-driving data and validating performance. A purpose-built, human-in-the-loop, customizable simulator capable of generating training data for different environmental scenarios and vehicle types for multiple drivers and crowdsourcing this information would accelerate research. 

Such a simulator could create a virtualized vehicle, synthetic sensor data, and present objectives encouraging ``good behavior.?? For example, a user could gain points from staying within lane markers, or by collecting coins placed along a trajectory, or by completing laps as quickly as possible with collision penalties. Collected data would aid behavior-cloning algorithms, without the cost of physical vehicles or driver pay. Due to the intentionally-variable nature of the simulator paired with human control, algorithms trained on synthetic data are more likely to learn invariant features, rather than features latent to the simulator design. The result will be improved algorithms capable of responding well to infrequent but impactful edge cases that other tools might miss, while the physical test platform will validate model performance in the real-world and help capture data for retraining model outputs for transfer learning.

\subsection{Training Deep Networks with Synthetic Data}
We will use synthetic images to train a deep learning model, and test the trained model on real-world images. This section explores training models using synthetic data and porting those models to the real-word. 

Neural network training is data-intensive, and typically involves collecting and manually annotating input prior to training. The collection labeling process is time consuming\cite{deng2009imagenet} and may require expert knowledge\cite{liu2015deep}. Labels may be difficult to identify even for humans\cite{zhou2016learning}. Generating high-quality, automatically labeled synthetic data helps to overcome these limitations. Common techniques include Domain Randomization (DR) and Domain Adaption (DA).

DR hypothesizes that a model trained on synthetic views augmented with random lighting conditions, backgrounds, and minor perturbations will generalize well to real-world conditions. DR's potential has been demonstrated in image-based tasks, including object detection\cite{tobin2017domain}, image segmentation\cite{ren2015faster} and object $6D$ pose estimation\cite{sundermeyer2018implicit, kehl2017ssd}. These methods render textured $3D$ models onto synthetic or real image backgrounds (e.g. MS COCO\cite{lin2014microsoft}) with varying brightness and noise levels. In this way, the domain gap between synthetic and realistic images can be reduced by increasing the generalizability of the trained model (small perturbations increase the likelihood of the model converging on invariant latent features).

Synthetic data also facilitates 3D vision tasks. For example, FlowNet3D\cite{liu2019flownet3d} is trained on a synthetic dataset (FlyingThings3D\cite{mayer2016large}) to learn scene flow from point clouds, and generalizes to real LIDAR scans captured in the KITTI dataset\cite{geiger2013vision}. Im2avatar\cite{sun2018im2avatar} reconstructs voxelized 3D models from synthetic 2D views from ShapeNet\cite{chang2015shapenet}, and the trained model produces convincing 3D models from realistic images of PASCAL3D+ dataset\cite{xiang2014beyond}. 

In contrast to DR, DA generates adapted images from synthetic images. Adapted images look similar to real-world images, and are used to train deep models. In general, models trained on DA data generalize well to real-world images. 

Generative Adversarial Networks (GANs)\cite{goodfellow2014generative} have been used to generate realistic data to train classifiers\cite{shrivastava2017learning}, 3D pose estimators\cite{bousmalis2017unsupervised} and grasping algorithms\cite{bousmalis2018using}. This work shows promising results, but the adapted images present unrealistic details and noise artifacts. 

We believe it should be possible to develop a simulator based on a game engine capable of generating meaningful data to inform Deep Learning self-driving models capable of real-world operation, with the added benefit of being able to crowdsource human control and trust the resulting input data as being ``clean.'' 

\section{The Gamified Digital Simulator}\label{gds} 
This section details the development of the Gamified Digital Simulator (GDS). The simulator blends the entertainment of a computer game with the utility of a scientific tool.

We first explore the overall application design, including layout, scenes, GameObjects, and scripts. We then describe the methodology for transforming the game into a tool for generating synthetic data for training a physical vehicle's line following model. Gamification allows non-experts to provide high-volume semi-supervised training data. This approach is unique relative to conventional simulation in that it provides a means of crowdsourcing data from goal-driven humans, expediting behavior cloning from the ``wisdom of the crowd.''
 
\subsection{Application Scenes}

\begin{figure*}[h]
 \centering
 \includegraphics[width=0.6\linewidth]{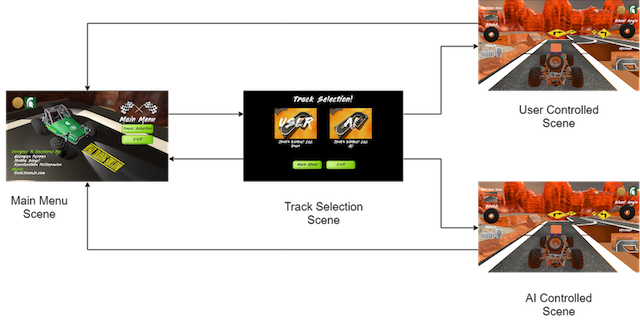}
 \caption{This figure shows the user-flow as the game scenes transition from the main screen through track selection and into the two operating modes (AI and human controlled)}
 \label{Scenes}
\end{figure*}

The GDS is built upon the Unity3D Game Engine to ease cross-platform deployment. The game consists of four scenes: a) Main Menu, b) Track Selection, c) User Input Mode and d) AI Mode. The Main Menu and Track Selection scenes contain canvases with UI elements helping the user transition between modes. From the Track Selection scene, a user selects one of two identical playable scenes. These scenes are controlled either by a human user (User Controlled Scene) or Artificial Intelligence (AI) (AI Controlled Scene). The AI Scene is utilizes in-game AI or an external script using TensorFlow\cite{tensorflow} or Keras\cite{keras} AI models, as described in Sections~\ref{in_game_ai} and~\ref{transferrability}.

The playable scenes consist of GameObjects including the virtual vehicle (``buggy''), track, surrounding environment, 2D Canvas elements, and cameras used to render the virtual in-car views. Scripts allow these objects to perform tasks including exporting virtual sample data recorded from human drivers, or executing or capturing the behavior of the in-game AI system. 

\subsection{Road Surface}

\begin{figure}[h]
 \centering
 \includegraphics[width=0.45\linewidth]{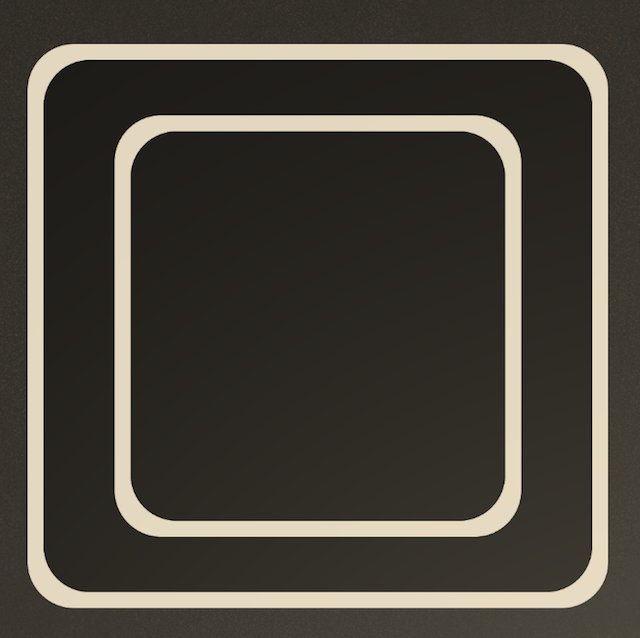}
 \caption{The initial test track was rectangular and designed as a simple test for the simulator and data collection system.}
 \label{Initial Rectangular Track}
\end{figure}

\begin{figure}[h]
 \centering
 \includegraphics[width=0.45\linewidth]{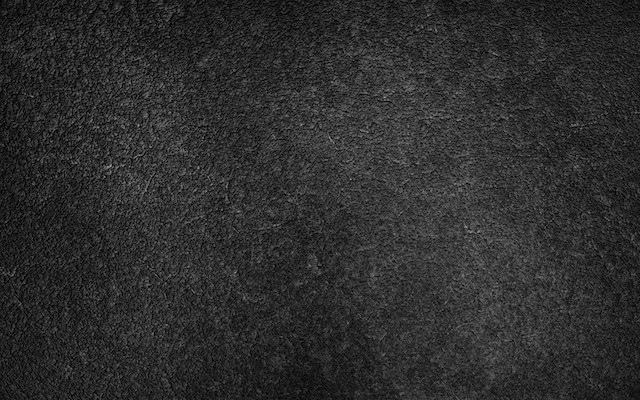}
 \caption{This asphalt texture was tessalated across the road surface in later versions of the simulator.}
 \label{Asphalt Texture}
\end{figure}

\begin{figure}[h]
 \centering
 \includegraphics[width=0.8\linewidth]{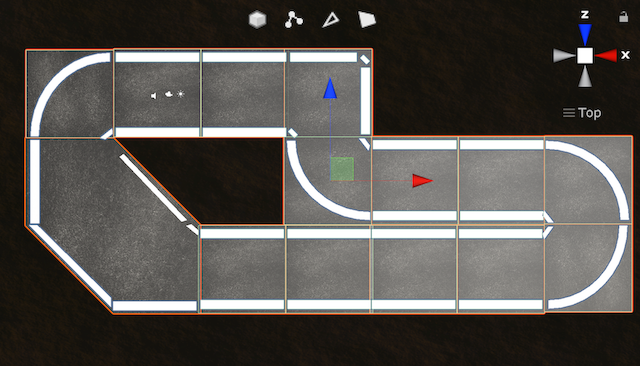}
 \caption{The track comprises adjacent modular components.}
 \label{Modular Track Components}
\end{figure}

The road surface GameObject simulates a test track (Figure~\ref{Initial Rectangular Track}). In order to create advanced track geometries, we developed modular track segments using Unity's cuboid elements for simple shapes, and ProBuilder for complex geometries (Figure~\ref{Modular Track Components}). Modules were given a realistic texture (Figure~\ref{Asphalt Texture}) that changes appearance based upon ambient lighting and camera angles, minimizing the likelihood that the neural network learns behaviors based on tessellated edge effects. 

To reduce model overfit, we developed a ``data augmentation'' script using Unity's lerp function to add Gaussian noise creating white speckles appearing over time. A second script manipulates the in-game lighting sources randomly so that vehicles repeating the same trajectory over time will varied training data. 

\subsection{Game Mechanics}
Gamification encourages players to abide by latent and overt rules to yield higher-quality, crowdsourcable data.

\begin{figure}[h]
 \centering
 \includegraphics[width=0.85\linewidth]{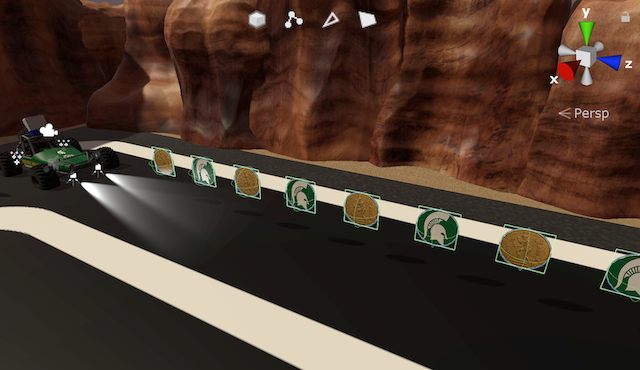}
 \caption{An early revision of the game featured collectable coins to incentive the driver to stay on the road surface. Each coin incremented a scoring counter. }
 \label{Coin Collection}
\end{figure}

On the rectangular track, we created GameObject coins within the lane markers. These elements gave users the opportunity to collect coins in order to enhance their score, creating intrinsic motivation for players to ``win'' (Figure~\ref{Coin Collection}). However, this approach allowed non-sequential collection, with users exiting the track boundaries and reentering without penalty. As a result, some generated data could contaminate learned models. 

\begin{figure}[h]
 \centering
 \includegraphics[width=0.85\linewidth]{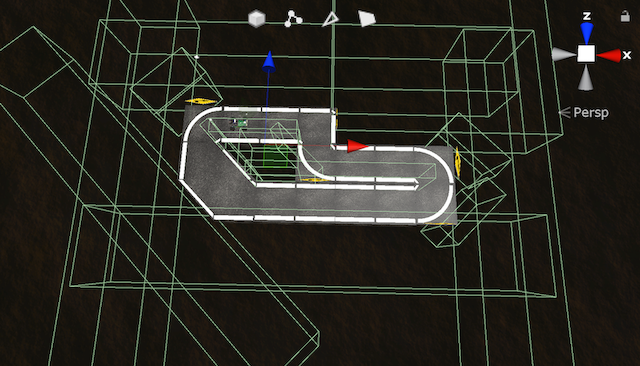}
 \caption{Rather than relying on users to follow implicit rules of the road, we developed an invisible collider system to force the buggy to stay on the road surface. }
 \label{Colliders}
\end{figure}

We subsequently developed a system of colliders (shown in green in Figure~\ref{Colliders}). Colliders invisible to the camera prevent buggy from traveling outside the white lane makers, ensuring that collected data are always within the lane markers, reducing contamination. 

\subsection{Environment}

\begin{figure*}[t!]
 \centering
 \begin{subfigure}[t]{0.27\linewidth}
 \centering
 \includegraphics[width=\linewidth]{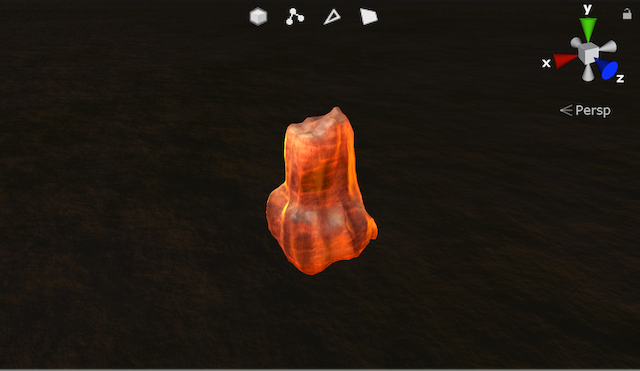}
 \caption{Rocky/Canyon 3D Model}
 \label{Rocky/Canyon 3D Model}
 \end{subfigure}
 ~
 \begin{subfigure}[t]{0.27\linewidth}
 \centering
 \includegraphics[width=\linewidth]{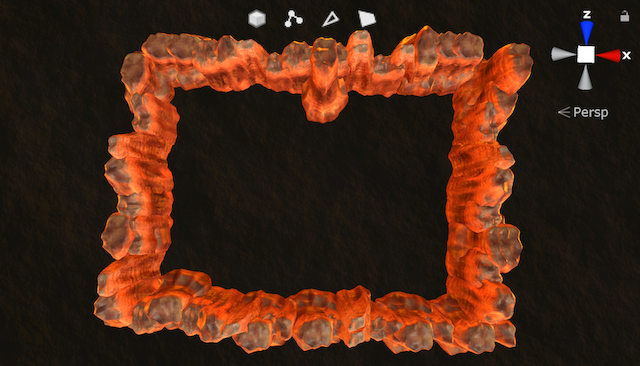}
 \caption{Mountain View}
 \label{Mountain}
 \end{subfigure}
 ~
 \begin{subfigure}[t]{0.27\linewidth}
 \centering
 \includegraphics[width=\linewidth]{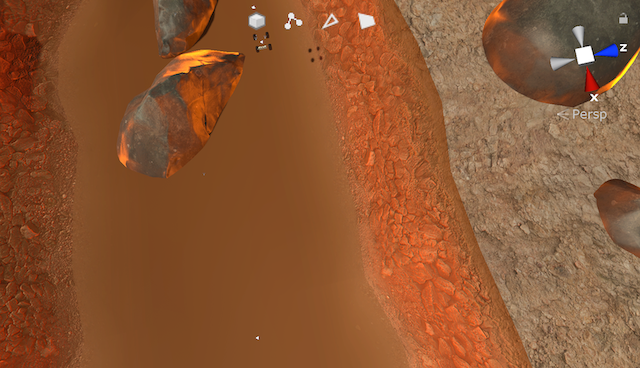}
 \caption{Lava stream terrain}
 \label{Lava stream terrain}
 \end{subfigure}
 ~ \\
 \begin{subfigure}[t]{0.27\linewidth}
 \centering
 \includegraphics[width=\linewidth]{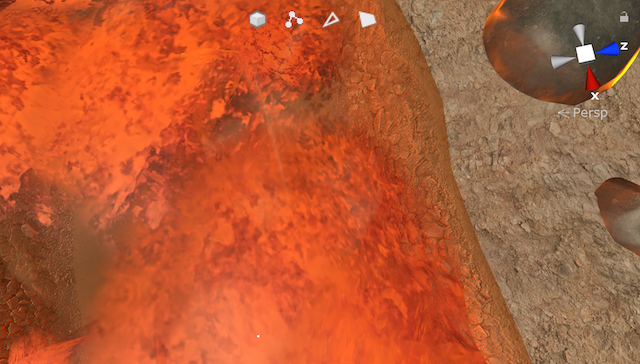}
 \caption{Lava particle effect in Play mode}
 \label{Lava Particle Effect}
 \end{subfigure}
 ~
 \begin{subfigure}[t]{0.27\linewidth}
 \centering
 \includegraphics[width=\linewidth]{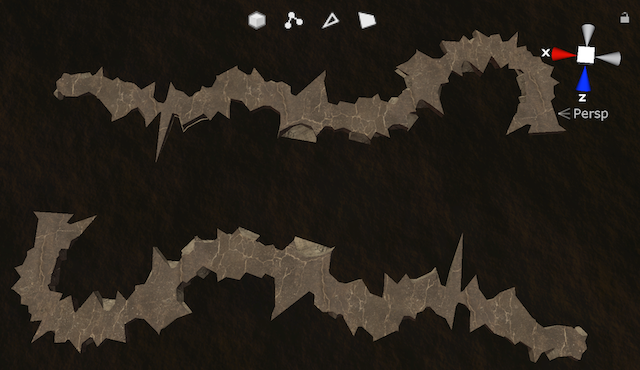}
 \caption{Ground of earthquake shift before earthquake happens}
 \label{Before Earthquake}
 \end{subfigure}
 ~
 \begin{subfigure}[t]{0.27\linewidth}
 \centering
 \includegraphics[width=\linewidth]{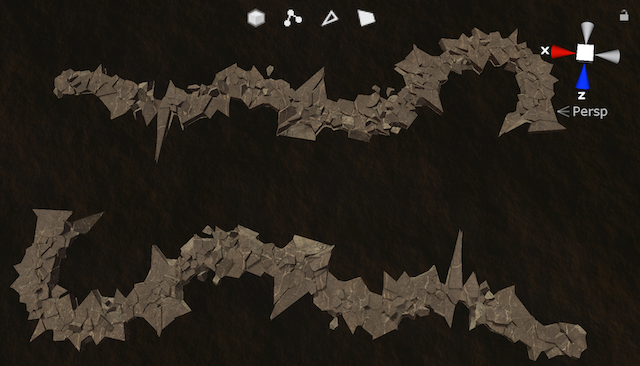}
 \caption{Ground of earthquake shift after earthquake happens}
 \label{After Earthquake}
 \end{subfigure}
 ~ \\
 \begin{subfigure}[t]{0.27\linewidth}
 \centering
 \includegraphics[width=\linewidth]{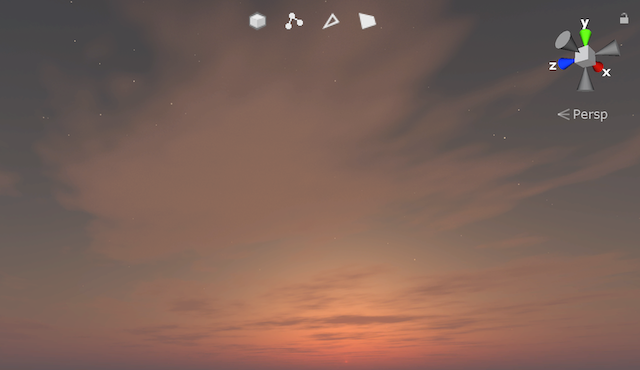}
 \caption{Night Skybox}
 \label{Skybox}
 \end{subfigure}
 ~
 \begin{subfigure}[t]{0.27\linewidth}
 \centering
 \includegraphics[width=\linewidth]{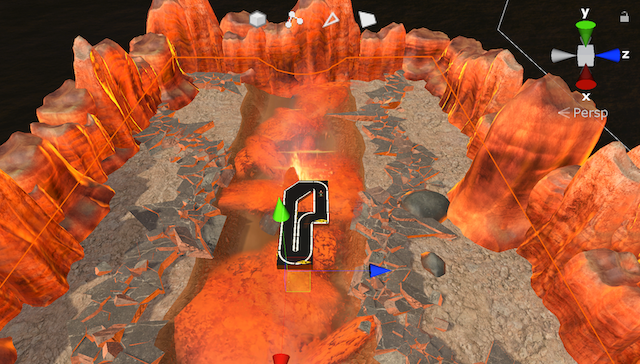}
 \caption{Combined environmental elements, showing relative scale of track}
 \label{Skybox}
 \end{subfigure}
 \caption{These figures show the elements comprising the simulated environment and surroundings. }
 \label{Environment and Surroundings}
\end{figure*}

The game environment has two purposes: to create a compelling user experience (UX) conducive to long play sessions, and to create dynamic conditions to avoid the neural network model fitting to environmental features. 

The game environment is a series of GameObjects and rendering lighting parameters. Objects include rocky surfaces and canyon models (Figure~\ref{Rocky/Canyon 3D Model}) scaled for the scene?s terrain size and arranged to create a rocky mountain view (Figure~\ref{Mountain}). We also included assets from the Unity Asset Store to excite players and create dynamic background images. These include a lava stream particle effect (Figure~\ref{Lava stream terrain} and Figure~\ref{Lava Particle Effect}) and two animated earthquake models (Figure~\ref{Before Earthquake}, Figure~\ref{After Earthquake}). Each asset was deployed within the scene to create the experience of driving near a volcano. The dynamic background provides ``noise'' in training images, helping ensure that the neural network fits to only the most-invariant features. Finally, in Rendering Options, a night Skybox with colors matching the overall aesthetics is applied (Figure~\ref{Skybox}).

The perceived hostility of the environment implicitly conveys the game mechanics - the buggy must not exit the track boundaries, or it will fall to its certain doom. This mechanic improves the ability for users to rapidly pick up and play the game. 


\subsection{Representative Virtual Vehicle}
The virtual vehicle is a multi-part 3D model. The 3D model was purchased from Unity Asset store and has customized colors, textures, and school logos. Atop the model, CAD from a physical vehicle's camera mount has been added to the assembly. 

\begin{figure}[h]
 \centering
 \includegraphics[width=0.80\linewidth]{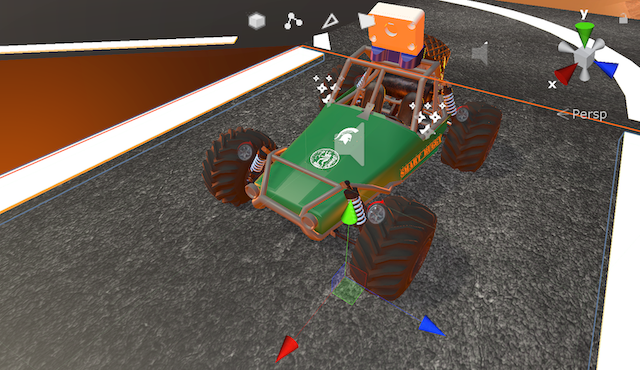}
 \caption{Each wheel of the buggy model has its own controller, running a high-fidelity vehicle physics model. This allowed us to precisely control steering angle, friction coefficient, and more. }
 \label{Buggy 3D model - Wheel Controllers}
\end{figure}

To make the buggy driveable, we created controllers for each wheel. Rather than using Unity's embedded wheel colliders, we purchased a plugin that simulates vehicle physics. Using this model, we could set parameters like steering angles, crossover speed, steer coefficient, Ackermann percentage, flip over behavior, forward and side slip thresholds, speed limiters and more. The ability to tune these parameters allowed us to tailor the in-game model to mimic the behavior of the physical vehicle without altering the model assembly. 

\subsection{Simulated Cameras}
The GDS uses a multi-camera system to simultaneously render the user view and export scientific data. The primary camera provides a third person perspective view to the user, while a second camera is placed at the front of the buggy and simulates the RC vehicle's real camera in terms of location, resolution and field of view (FoV) characteristics. This camera generates the synthetic images used for training the neural network model and is calibratable in software. A third camera is placed high above the vehicle, facing downwards. This camera provides an orthographic projection along with a RenderTexture and creates a mini-map for the user.

Two additional cameras behave akin to SONAR or LIDAR and calculate distance to nearby objects using Unity?s Raycasting functionality. These cameras support in-game ``artificial intelligence'' used to generate synthetic training data without human assistance (Section~\ref{in_game_ai}). 

The location of the cameras is linked relative to the buggy's body. Each camera has differing abilities to ``see'' certain GameObjects. This behavior is controlled using Unity's Layer functionality. For example, the primary camera sees turn indicators directing the user, but the synthetic forward imaging camera ignores these signs when generating training data. In the coin demo, the user could see the coins, but the synthetic camera treated those objects as being invisible. 

\subsection{Exported Data}
Exporting accurate data is critical. We developed a script attached to the buggy and use this to sample data at regular intervals, capturing a timestamp, the buggy's speed, and local rotation angle of the front wheels. This information is logged to a CSV in the same format used by the physical platform. Another script captures images from the synthetic front-facing camera to JPG at each timestep to correlate the steering angle and velocity with a particular image. The CSV file and the images are stored within the runtime-accessible StreamingAssets folder. 

\subsection{Simulator Reconfigurability}

\begin{figure}[h]
 \centering
 \includegraphics[width=0.95\linewidth]{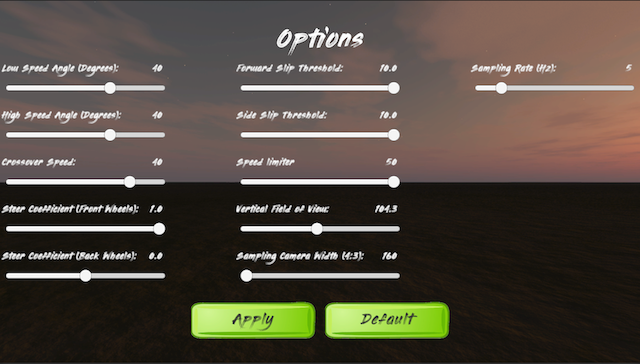}
 \caption{When the simulator is connected to a multi-display host, the second display allows users to adjust options in real-time. This greatly speeds up tuning the buggy model to match the physical model, and helps make the simulator more extensible to other vehicle types. }
 \label{Options Screen}
\end{figure}

Adjusting GameObject or plugin settings is a time-consuming process. In order to simulate the configuration of the physical vehicle, we needed to conduct and iterate upon multiple tests to converge on parameters closely approximating the physical vehicle. To speed this process up and to allow for the simulator's applicability to other vehicle types, we developed an \textit{Options.pref} file. Important vehicle and simulator options can be changed by editing the file. Configurable options include:

\begin{enumerate}

\item{\textbf{Low speed steering angle} - the absolute value of the maximum angle at which the center of the steered wheels is maximum when the vehicle is in low-speed mode}
\item{\textbf{High speed steering angle} - same as above, but in high-speed mode}
\item{\textbf{Crossover speed} - the speed at which the vehicle changes from low- to high-speed mode}
\item{\textbf{Steer coefficient (front wheels)} - the steering multiplier between the steering angle and the actual wheel movement}
\item{\textbf{Steer coefficient (back wheels)} - same as above. Can be negative for high-speed lane changes (translation without rotation)}
\item{\textbf{Forward slip threshold} - slip limit for transition from static to sliding friction when accelerating/braking}
\item{\textbf{Side slip threshold} - same as above, but for steering}
\item{\textbf{Speed limiter} - maximum vehicle speed allowable (also reduces available power to accelerate)}
\item{\textbf{Vertical Field of View} - in degrees, to match physical vehicle camera}
\item{\textbf{Sampling Camera Width} - ratio of width to height of captured image}
\item{\textbf{Sampling Rate} - rate, in Hz, of capture of JPG images and logging to CSV file}
\end{enumerate}

In addition to the Options file, another screen appears on the second monitor in multi-display environments. This screen contains sliders with all the options and lets the user to make changes visually (Figure~\ref{Options Screen}).

\begin{figure}[h]
 \centering
 \includegraphics[width=0.85\linewidth]{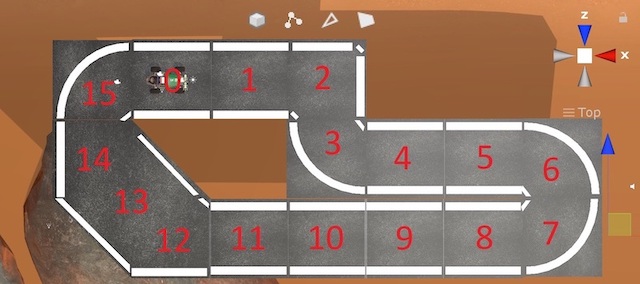}
 \caption{This figure shows the allowable starting position options selectable in the PosRot.spawn file. Each number refers to the center of the tile, while a starting angle can also be passed as a parameter to the game engine. }
 \label{Position}
\end{figure}

We also created a \textit{PosRot.spawn} file to set the starting coordinates of the buggy (Figure~\ref{Position}). This helps test humans and AI alike under complicated scenarios (e.g. starting immediately in front of a right turn where the horizontal line is exactly perpendicular to the vehicle, or starting perpendicular to the lane markers).

\subsection{2D Canvas Elements}
\begin{figure}[h]
 \centering
 \includegraphics[width=0.85\linewidth]{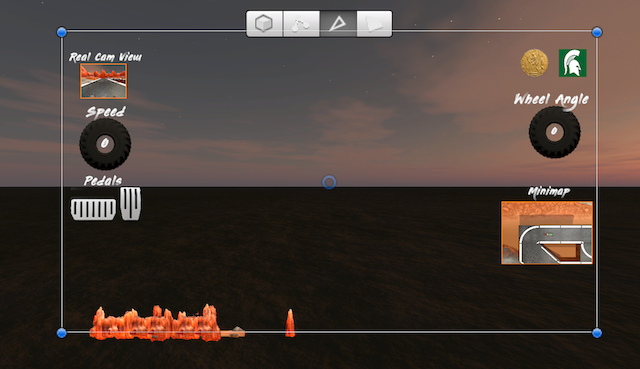}
 \caption{2D Canvas Elements include speed and steering displays, a minimap, the forward camera view, and a minimap.}
 \label{2D Canvas Elements}
\end{figure}

2D GameObjects display elements show the buggy's speed and wheel angle to the user in real time. There are also two RenderTexture elements, one displaying a preview of the synthetic image being captured (the ``real cam'' view) and one showing the track from above (the ``minimap''). Throttle and brake status indicators turn green when a user presses the associated buttons on the game controller (Figure~\ref{2D Canvas Elements}). This feature helps users understand the controls, making it easier for newcomers to drive well and capture useful training data quickly. 

\subsection{Unity C\# and Python Bridge}\label{python_bridge}
We test models in the virtual environment before porting them to a physical vehicle. Deep learning frameworks commonly run in Python environments, whereas Unity supports C\# and JavaScript.

We developed a bridge between Unity and Python in the form of an \textit{AI.input} file. This file, located in the StreamingAssets folder, contains three values: commanded steering angle, commanded velocity, and AI mode (in-game AI -- using the cameras described in Section~\ref{in_game_ai}, or external AI from Section~\ref{data_collection}, where the commanded steering and velocity values are read from the file at every loop). 

This approach allows near-realtime control from an external model. Python scripts monitor the StreamingAssets folder for a new image, processes this image to determine a steering angle and velocity, and updates the file with these new values. The simulator uses these values to control the steering and velocity. The loop takes a few milliseconds from output image to prediction to input values, so the impact of latency is minimal.

Using the PyGame library\cite{pygame}, we also ``pass through'' non-zero gamepad values, allowing for semi-supervised vehicle control (AI with human overrides). This approach allows us to test models in-game easily, and helps us to ``unstick'' vehicles to observe a model's continuation after encountering a complex scenario. We allow the vehicle to drive itself until the user engages with the joystick, which supersedes the external model's control. Upon releasing the controller, the external AI resumes control.  

\subsection{User Controlled Input and In-game AI Modes}\label{in_game_ai}
There are two game modes: User Control and AI Control. 

In the User Controlled scene, users interface with gamepads' analog joysticks to control the vehicle. Inputs can be custom-mapped when the game launches. 

\begin{figure}[h]
 \centering
 \includegraphics[width=0.85\linewidth]{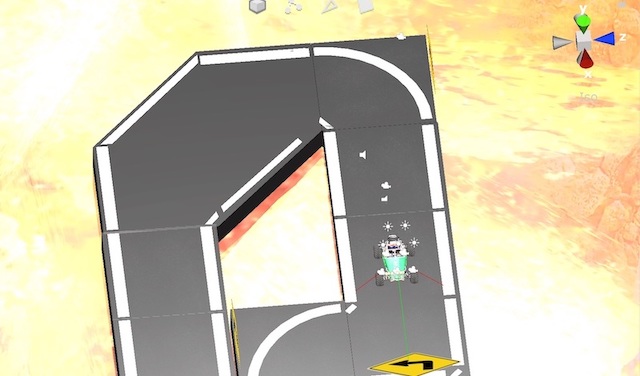}
 \caption{This figure shows the cameras used to measure the distance to the environment via raycasting. Here, the green line projected from the front camera indicates the longest unoccupied distance to an object, while the two red lines indicate the distance measures for each of the two side cameras. The in-game AI aims to keep these two distances roughly equal to center the buggy in the lane. }
 \label{Raycasting}
\end{figure}

In the AI Controlled scene, the buggy is controlled by in-game AI or an external model (Section~\ref{python_bridge}). 

In-game AI provides a means of capturing simulated data without user input. This AI uses a three-camera system (the front camera [the same used to capture the synthetic view image] and two side cameras rotated $\pm60^{\circ}$ relative to the y-axis) as input. Each camera calculates the distance between its position and the invisible track-border colliders. 

The camera orientations and ``invisible'' distance measurements are represented in Figure~\ref{Raycasting} (the green line represents the longest clear distance, and two red lines indicate more-obstructed pathways). The car moves in the direction of the longest free distance. If the longest distance comes from the front camera, the buggy moves straight ahead. If it comes from a side cameras, then it centers itself in the available space. When the front distance falls under a braking threshold, the buggy slows in advance of a turn. 

Unity is unable to simulate joystick input. Instead, we use the same bridge from Section~\ref{python_bridge} to both write and read output for controlling the buggy. This in-game AI method generated trustable unsupervised training data for the Python deep learning network described in Section~\ref{data_collection}. The in-game AI model is like learning to ride a bicycle with training wheels - and the learned deep learning network is the model balancing on two wheels once it's gotten enough practice.

\section{Integrating GDS with An End-To-End Training Platform}
The GDS is part of an end-to-end training system for self driving. Our system comprises a physical platform (used for model validation and to inspire the physics and sensor parameters for the virtualized vehicle), a physical training environment (duplicated in the simulated world), and the GDS (emulating the vehicle, its sensors, and its environment while encouraging desirable behavior). Each element is detailed in the following subsections. 

\subsection{A Physical Self-Driving Test Platform}
Self-driving models are designed for physical vehicles. We therefore had to create a physical vehicle platform and environment duplicating the GDS world to validate model performance. 

We considered the TurtleBot\cite{turtlebot} and DonkeyCar\cite{donkeycar} platforms, as both are low-cost development systems. The TurtleBot natively supports the ROS middleware and Gazebo simulation tool, however the kinematics of the differential-drive TurtleBot do not mirror conventional cars. The DonkeyCar offers Ackermann steering and a more powerful powertrain with higher top speed, better mirroring passenger vehicles. However, the DonkeyCar did not run ROS, and therefore would be less useful as an extensible research and development platform.

We therefore developed a self-driving platform using hardware similar to the DonkeyCar, and a software framework similar to that of the TurtleBot ? a $1$/$10^{th}$ scale radio-controlled car chassis with Ackermann steering, running ROS. 

Computing is provided by a Raspberry Pi 3B+, while a Navio2\cite{navio2} provides an onboard Inertial Measurement Unit (IMU) and I/O for RC radios, servos and motor controllers. Additional sensor input is provided by a Raspberry Pi camera with 130 degree field of view and IR filter (to improve daytime performance), and optionally a 360-degree planar YDLIDAR X4\cite{rp_lidar} to measure radial distances. The platform is connected to a Logitech F710 dual analog joystick through the USB port for human control. 

The Raspberry Pi runs the Raspian Stretch OS with realtime kernel as provided by Emlid, the maker of Navio. At boot, the OS launches the $Ardupilot$\cite{ardupilot} service, $mavros$\cite{mavros}, and the $joy$ node. If used, the $rplidar$ node is loaded. The user then launches one of two Python nodes via SSH: a $teleop$ note, which uses the joystick to control the car and logs images, IMU sensor data, joystick inputs, and servo and motor outputs to an onboard SD card at $10$Hz, or a $model$ node, which uses one or more camera images and a pretrained model to command the steering servo to follow line markers using a pretrained neural network. In this mode, the user manually controls the throttle using the F710. Motor and servo commands take the form of a pulse width command ranging from $1000$-$2000$uS, published to the $/mavros/rc/override$ topic. 

\begin{figure}[t!]
 \centering
 \includegraphics[width=0.8\linewidth]{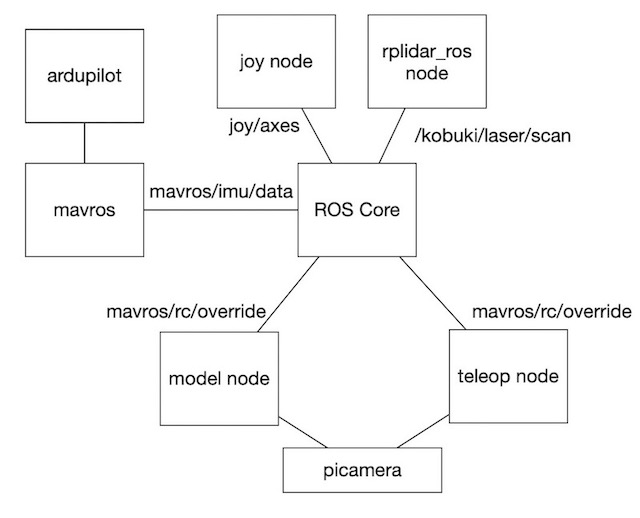}
 \caption{A series of ROS nodes communicate with the ROS Core in order to exchange telemetry, sensor data, and control information. }\label{ros}
\end{figure}

When the $teleop$ mode is started, the IMU's accelerometer and controller data are captured by ROS subscribers and written to .CSV file, along with the $160x120$ RGB .JPG image captured from $picamera$ at the same time step. The accelerometer is recorded for use in future work. An overview of ROS architecture, including nodes and topics, appears in Figure~\ref{ros}.

\begin{figure}[t!]
 \centering
 \includegraphics[width=0.8\linewidth]{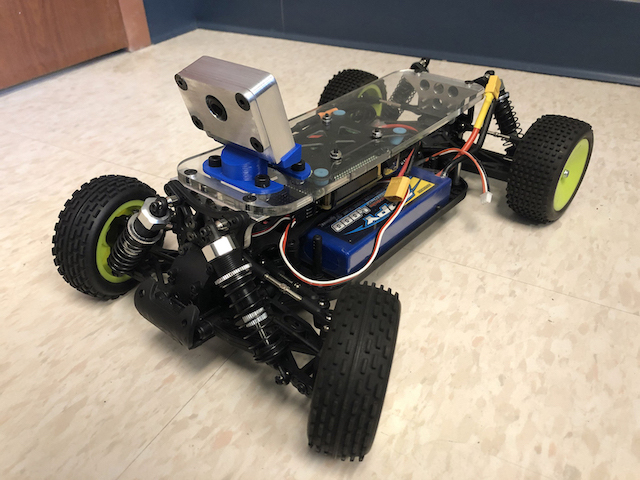}
 \caption{This 1/10th scale buggy features a Raspberry Pi 3B+, Navio2 interface board, Logitech F710 USB dongle, and a Raspberry Pi camera. Not pictured: LIDAR. }\label{vehicle_photo}
\end{figure}

The hardware platform is based on a $1$/$10^{th}$ scale Short Course Truck (SCT) from HobbyKing. The platform is four-wheel drive and has a brushless motor capable of over $20$kph. The Pi is vibrationally-isolated on an acrylic plate, reducing mechanical noise and providing crash protection. The camera is mounted atop the same acrylic plate and protected by an aluminum enclosure to minimize damage during collisions. The camera is mounted to a 3D printed bracket, the angle of which was set experimentally to provide an appropriate field of view for line detection. The LIDAR, if used, is mounted to this same plate using standoffs to raise the height above the camera enclosure. The vehicle platform is shown in Figure~\ref{vehicle_photo}. 

This platform utilizes widely-adopted ROS tools and is small, inexpensive, robust, and easy to repair, helping to democratize automated vehicle development. 

\subsection{Modular Training Environment}
\begin{figure*}[t!]
 \centering
 \begin{subfigure}[t]{0.32\textwidth}
 \centering
 \includegraphics[height=1.1in]{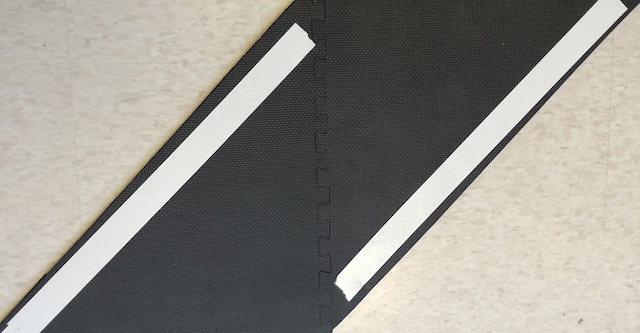}
 \caption{This monomer, made up of two half-tiles, is a high-speed and very narrow ``lane change'' piece.}
 \end{subfigure}%
	~
 \begin{subfigure}[t]{0.32\textwidth}
 \centering
 \includegraphics[height=1.1in]{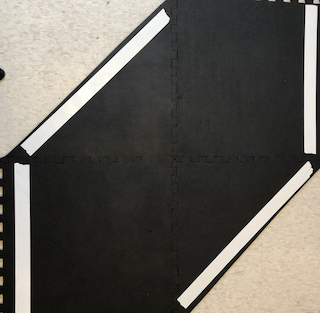}
 \caption{This monomer, made up of two half-tiles and two full tiles, is a high-speed and very wide ``lane change'' piece. It is difficult to navigate as only one lane marker is visible at a time (or none, if the car is perfectly centered).}
 \end{subfigure}%
	~
 \begin{subfigure}[t]{0.32\textwidth}
 \centering
 \includegraphics[height=1.1in]{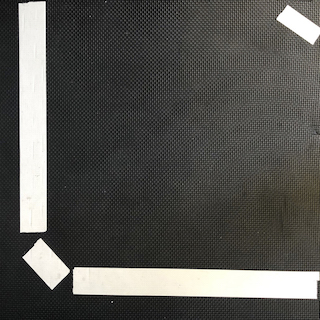}
 \caption{This monomer, made up of one tile, is a very sharp 90-degree turn. When approached head-on, there are only two small angled pieces of tape to differentiate left from right turns (the vertical and horizontal lines dominate the field of view).}
 \end{subfigure}%
	~ \\
 \begin{subfigure}[t]{0.45\textwidth}
 \centering
 \includegraphics[height=1.1in]{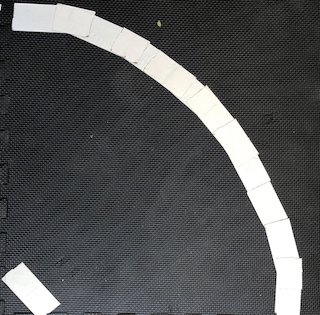}
 \caption{This monomer is a sweeping turn made up of one tile, and is fairly easy for the vehicle to follow. However, the turn radius is almost exactly the maximum allowed by the car's steering geometry.}
 \end{subfigure}%
	~
 \begin{subfigure}[t]{0.45\textwidth}
 \centering
 \includegraphics[height=1.1in]{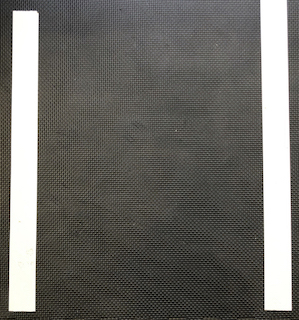}
 \caption{This monomer is a single-tile straightaway, slightly wider than the width of the buggy.}
 \end{subfigure}%
 \caption{These monomers are made of $24$'' square EVA foam interlocking gym tiles and combine to form a range of track configurations with varying complexity. }\label{monomers}
\end{figure*}

We designed a test track using reconfigurable ``monomer'' building blocks to create a repeatable, reconfigurable testing environment. We created track elements using low-cost $\frac{1}{2}$'' thick EVA foam gym tiles. The library of components included tight turns, squared and rounded turns, sweeping turns, straightaways, and lane changes. See samples of each tile in Figure~\ref{monomers}.

Straightaways and rounded (fixed-width) corners provide the simplest features for classification; the wide, sweeping and hard-right-angle corners provide challenging markings. 

\begin{figure*}[t!]
 \centering
 \begin{subfigure}[t]{0.45\textwidth}
 \centering
 \includegraphics[height=1.6in]{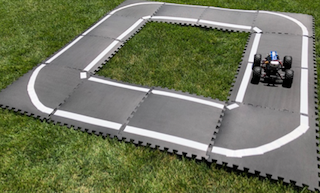}
 \caption{This outdoor track layout is similar to the original rectangular surface used in the coin-collecting game.}
 \end{subfigure}%
	~
 \begin{subfigure}[t]{0.45\textwidth}
 \centering
 \includegraphics[height=1.6in]{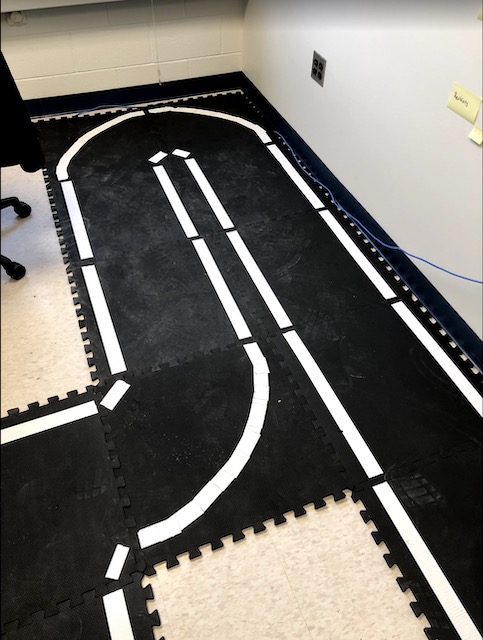}
 \caption{This track mimics the track geometry used in the simulator with invisible colliders.}
 \end{subfigure}%
 \caption{We tested multiple physical tracks both matching the layouts in-game and of entirely new designs.}
\label{track_layouts}
\end{figure*}

The reconfigurable track is quick to setup and modular compared with placing tape directly on the ground. It also helps to standardize visual indicators, similar to how lane dividers have fixed dimensions depending on local laws. Inexpensive track components can be stored easily, making this approach suitable to budget-constrained organizations. Sample track layouts appear in Figure~\ref{track_layouts}. 

\subsection{Integration with Gamified Digital Simulator}\label{game_design}
Simulation affords researchers low-cost, high-speed data collection across a range of environments. We use the GDS to parallelize data collection for lane keeping algorithms from multiple users without the space, cost, or setup requirements associated with conventional vehicles. We aim for the physical vehicle to ``learn'' to drive by camera using GDS inputs. 



In Section~\ref{gds}, we describe the creation of an in-game proxy for the physical platform. The virtual car is rigged to mirror the real vehicle's physics, and driven using the same F710 joystick. The simulator roughly matches the friction coefficient, speed, and steering sensitivity between the vehicles, with numeric calibration where data were readily quantifiable (e.g. field of view for the camera), and by feel otherwise. 



The virtual car saves images to disk in the same format as the physical vehicle to maximize data interoperability, and exports throttle position, brake position, and steering angle to a CSV. 



\section{Data Collection and Algorithm Implementation}\label{data_collection}
To prove the viability of the simulator as a training tool, we designed an experiment to collect line-following data in the simulated environment for training a self-driving Convolutional Neural Network (CNN). We chose to use a CNN to identify road marker lines and complex features such as tight and sweeping corners that challenge traditional Hough Transforms.

We generated training data both from ``wisdom of the crowd'' (human drivers) and from ``optimal AI'' (steering and velocity based on logical rules and perfect situational information). We then attempted to repurpose the resulting model, without modification, to the physical domain. The process of collecting, augmenting, building, and validating a model on these data is described below. 

We first collected data in the simulated environment by manually driving laps in 10-minute batches. As described in Section~\ref{game_design}, the simulated camera images, steering angles, and throttle positions were written to file at $30$Hz. These data were filtered so that samples with zero velocity, negative throttle (braking/reverse), or steering outside the control limits (indicating a collision with a collider) were ignored. This prevented the algorithm from learning from ?near crash? situations.


We initially planned to augment these data with multiple sets of semi-synthetic training data to increase input diversity, each with randomly-added Gaussian blur, Gaussian noise, contrast enhancements, per-channel color shifts, and small-scale skewing, translation, and rotation. In practice, tuning augmentation parameters proved a difficult balance between model generalizability and eroding critical features. Instead, we used the in-game ``optimal AI'' (AI with complete, noiseless sensor measurements and well characterized rules as described in Section~\ref{in_game_ai}) to generate additional ground-truth data from the simulator and ultimately used only symmetric augmentation (image and steering angle mirroring to ensure left/right turn balance). In this sense, the AI augments the human data to prevent overfit - and vise-versa. 

The final data were approximately half human-driven and half controlled by AI with noiseless information. We collected $224,293$ images, almost $450,000$ images after symmetry augmentation. Each image was then post-processed to extract informative features. 

We developed an image preprocessing pipeline in OpenCV\cite{opencv} capable of:
\vbox{%
\begin{enumerate}
\item{Correcting the image for camera lens properties}
\item{Conducting a perspective transform to convert to a top-down view}
\item{Conversion from RGB to HSL color space}
\item{Gaussian blurring the image}
\item{Masking the image to particular ranges of white and yellow}
\item{Greyscaling the image}
\item{Conducting Canny edge detection}
\item{Masking the image to a polygonal region of interest}
\item{Fitting lines using a Hough transform}
\item{Filtering out lines with slopes outside a particular range}
\item{Grouping lines by slope (left or right lines)}
\item{Fitting a best-fit line to the left and/or right side using linear regression}
\item{Creating an image of the best-fit lines}
\item{Blending the (best-fit) line(s) with the edge image, greyscale image, or RGB image}
\end{enumerate}}

Using a sample CNN, we permuted operations and identified a subset as being critical to predictor accuracy. Final preprocessing steps included:

\begin{enumerate}
 \item{Conversion from RGB to HSL color space}
 \item{HSL masking to allow only white and yellow regions to pass through into a binary (black/white) image}
 \item{Gaussian blurring}
 \item{Masking the image to a region including only the road surface and none of the vehicle or environs}
 \end{enumerate}

\begin{figure*}[t!]
 \centering
 \begin{subfigure}[t]{0.5\textwidth}
 \centering
 \includegraphics[height=1.15in]{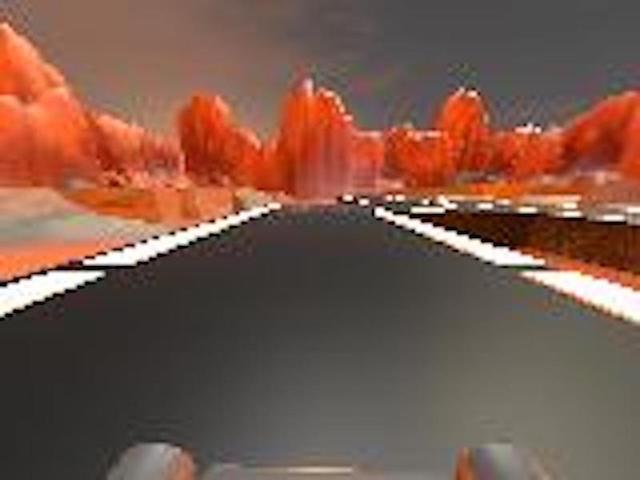}
 \caption{This is a sample RGB input image from the simulated environment. }
 \end{subfigure}%
 ~ 
 \begin{subfigure}[t]{0.5\textwidth}
 \centering
 \includegraphics[height=1.15in]{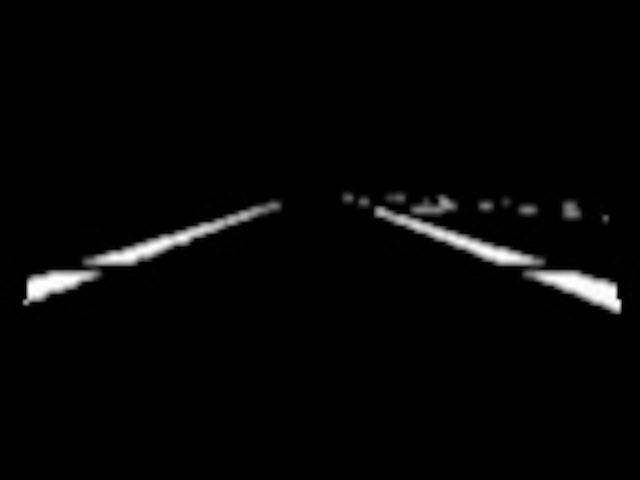}
		\caption{This image shows the preprocessed camera input (B\&W) for the simulated environment. The lines stand out relative to the rest of the image, improving classifier robustness. }
 \end{subfigure}
 ~\\
 \begin{subfigure}[t]{0.5\textwidth}
 \centering
 \includegraphics[height=1.15in]{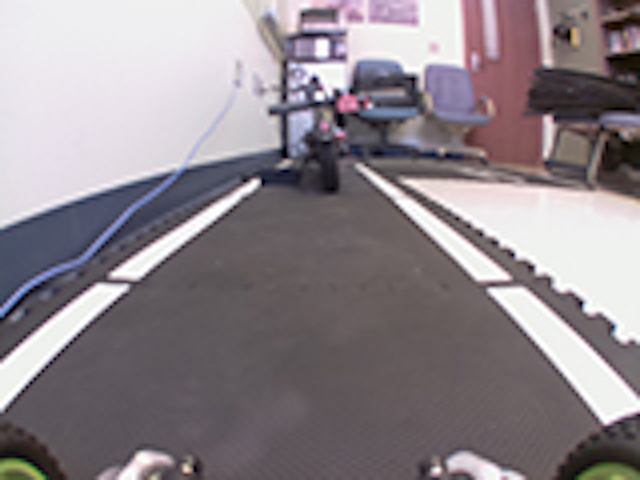}
 \caption{This is a sample RGB input image from the Raspberry Pi camera on the physical vehicle.}
 \end{subfigure}%
 ~ 
 \begin{subfigure}[t]{0.5\textwidth}
 \centering
 \includegraphics[height=1.15in]{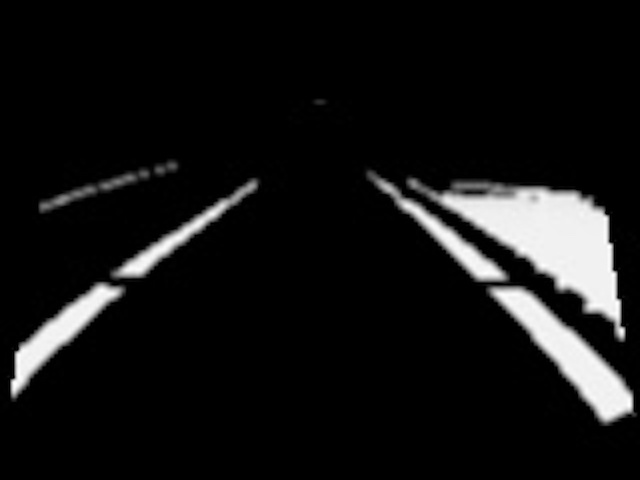}
		\caption{This image shows the preprocessed camera input (B\&W) for the real environment. The extracted features are similar between the simulated and physical world, though there is more noise and the white floor is retained as a potential ``line feature.'' In many road environments, nearby features with the exception of curbs would not be light white. }
 \end{subfigure}
 \caption{Preprocessing images is an efficient process that improves model transferrability between the simulated and physical world.}\label{images_processed}
\end{figure*}

This pipeline improved predictor robustness for varied lighting conditions with minimal increase in computation time compared to processing and classifying raw RGB images. Example input and output images from the real and simulated camera appear in Figure~\ref{images_processed}.

We additionally tested camera calibration and perspective transformation, but found these operations to add little performance relative to their computational complexity. The predicted steering angles and control loop update sufficiently quickly that over- or under-steering is easily addressed. We did implement a maximum steering slew rate in the predictor to prevent the vehicle from changing steering direction abruptly while minimizing oscillation. 

We trained more than $25$ CNN variants in Keras\cite{keras} using only simulated data, each with differing layers, color channels and image sizes, batching, pooling, normalization, dropout operations, and learning rates. For each convergent model, we recorded outsample mean-squared error (MSE). We saved the best model and stopped training when the validation loss had not decreased more than $0.1$ across the previous $200$ epochs. In practice, this meant training for approximately $400$ epochs. In all cases, testing and validation loss decreased alongside each other for the entirety of training, indicating that the model did not overfit. 
 
\begin{figure*}[t!]
 \centering
 \begin{subfigure}[t]{0.5\textwidth}
 \centering
 \includegraphics[height=5.0in]{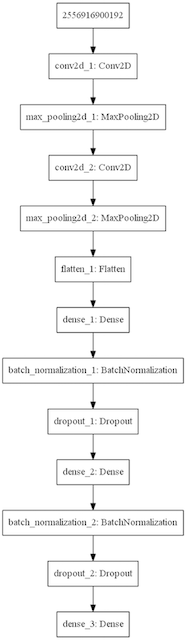}
 \caption{This 2D Convolutional Neural Network is the single-image model, taking a greyscale $160$x$120$ image as input and outputting a single (float) angle.}\label{image_model}
 \end{subfigure}%
 ~ 
 \begin{subfigure}[t]{0.5\textwidth}
 \centering
 \includegraphics[height=5.0in]{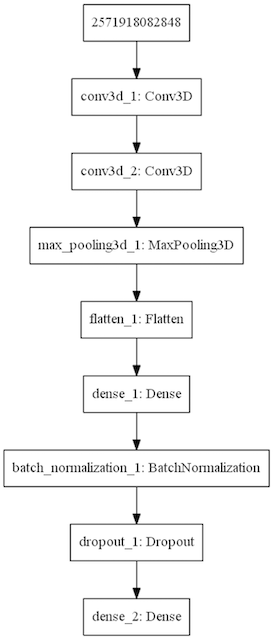}
		\caption{This 3D Convolutional Neural Network is the single-image model, taking three greyscale $160$x$120$ images as input and outputting a single (float) angle.}\label{sequence_model}
 \end{subfigure}
 \caption{The two models used to predict steering angle from greyscale images are both convolutional neural networks (CNN's) - 2D for the single-image case, and 3D for the image-sequence case.}\label{cnns}
\end{figure*}

From these results, we selected two CNN's with the best outsample performance: one using a single input image, and one using a sequence of three images (current and the two preceding images) to provide time history and context. The final models are shown in Figure~\ref{image_model}~and~\ref{sequence_model}. 

\begin{figure*}[t!]
 \centering
 \begin{subfigure}[t]{0.45\textwidth}
 \centering
 \includegraphics[height=2.1in]{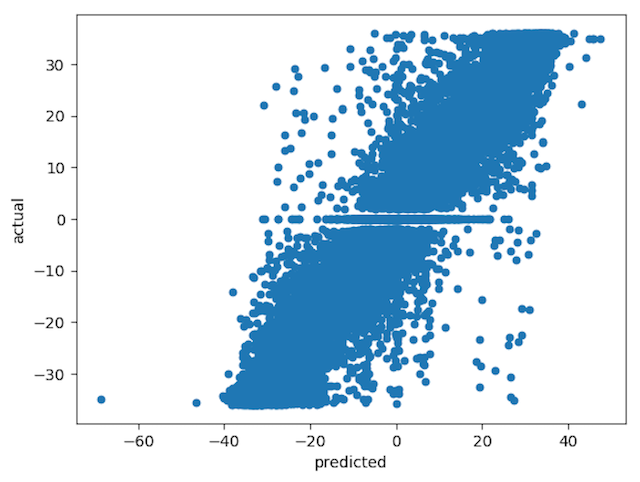}
 \caption{This figure shows the predicted versus ground-truth steering angles for the single-image model, with a reasonably strong diagonal component despite few outlying values. }
 \end{subfigure}%
 ~ 
 \begin{subfigure}[t]{0.45\textwidth}
 \centering
 \includegraphics[height=2.1in]{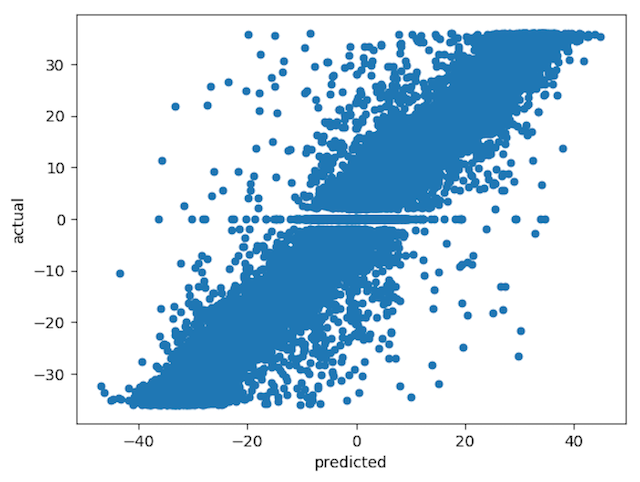}
 \caption{This figure shows the predicted versus ground-truth steering angles for the multiple-image model, with a strong diagonal clustering.}
 \end{subfigure}
 \caption{Comparing the model performance for the single- and multi-image predictor, using in-game images captured at approximately $30$ frames per second.}\label{compare_diagonals}
\end{figure*}

A comparison of the predictive performance of the single-image and multi-image model for the (simulated) validation set appears in Figure~\ref{compare_diagonals}. These plots compare the predicted steering angle to the ground-truth steering angle, with a $1:1$ slope indicating perfect fit. 

\section{Model Testing and Cross-Domain Transferrability}\label{transferrability}
We tested the trained model in both virtual and physical environments to establish qualitative performance metrics and provide commentary on domain transferrability without retraining.

\subsection{Model Validation (Simulated)}
We first tested the model in the simulator's environment, using Keras to process the latest output images from the virtual camera. The neural network monitored the image output directory, running each new image through a pretrained model to predict the steering angle. This steering angle and a constant throttle value were converted to simulated joystick values and written to an input file monitored by the simulator. The simulator updated the vehicle control with these inputs at $30$Hz, so the delay between the image creation, processing, and control input was $0.06$ seconds or better. This process is described in Section~\ref{python_bridge}.

The simulated vehicle was able to reliably navigate along straightaways and fixed-radius turns. It also correctly identified the directionality of tight and sweeping corners with high accuracy, though the classifier struggled particularly with the hard-right-angle turns, suggesting the model identifies turns by looking for curved segments rather than by tile corner geometry. For some initial seeds of starting position, angle, and lighting, the simulated vehicle could complete $>20$ laps without incident. For other seeds, the vehicle would interact with the invisible colliders and ``ping-pong'' against the walls (directional trends were correct, but tight-radius turns and narrow lanes left little room for error).  In these cases, human intervention unstuck the vehicle and the buggy would resume driving in the center of the lane. 

The simulator was able to train a model capable of performing in a virtual environment, so we transferred the model to the physical vehicle without retraining. The next section qualitatively describes the physical vehicle's performance. 

\subsection{Model Transferrability to Physical Platform}
We ported the pretrained model to the physical vehicle unchanged, but did alter the HSL lightness range for OpenCV's white mask and changed the polygonal mask region to block out the physical buggy's suspension (appearing at different pixel locations than the virtual buggy), and scaled the predicted output angle (converting degrees to microsecond servo pulses). There was no camera calibration and no perspective transformation despite the different camera tilt angles and lens types. 

The buggy was able to follow straight lines and sweeping corners using the unaltered single- and multi-image models, with the vehicle repeatedly completing several laps before incident. It was not necessary to use any real images to retrain the model's output (though retraining the last layer or entire model with some real images may improve robustness). Both the real car and simulator control loops operate at low loop rates ($8$-$30$Hz) and speeds (~$10$kph), so the classifiers? predicted steering angles cause each vehicle to behave as though being operated with a ``bang-bang'' controller (``left-right'') rather than a nuanced PID controller. Though line-following appears ``jerky,?? the low speed also means that small disparities between the calibration and sensitivity of the virtual and physical vehicle?s steering response minimally impact line-following performance. 

We also qualitatively evaluated the single- and multi-image models relative to their performance in the simulated environment. In practice, the model relying on the single image worked most robustly within the simulated environment. This is because the images for training were captured at a constant $30$ frames per second, but the vehicle speed varied throughout these frames. Because the 3D convolution considers multiple frames at fixed time intervals, there is significant velocity dependence. The desktop was able to both process and run the trained model at a consistent $30$ frames per second, including preprocessing, classification, polling for override events from the joystick, and writing the output file, making the single-image model perform well and making the impact of reduced angular accuracy relative to the image-sequence model insignificant as the time-delta between control inputs was only $0.03$ seconds.

In the physical world, where the buggy speed and frame capture and processing are slower, the vehicle's velocity variation is a smaller percentage of the mean velocity and computational complexity matters more. As a result, the image sequence provides significantly better performance for the physical vehicle as it anticipates upcoming turns without the complication of high inter-frame velocity variation. The physical vehicle performed laps consistently with the image-sequence model, though it still struggled with right-angle turns similar to the simulator (both with ``optimal AI'' and the neural network approach, suggesting this is simply a more complicated problem). 

These results show successful model transferrability from the simulated to physical domain without retraining or recalibration. The gamified driving simulator and low-cost physical platform demonstrate an effective end-to-end solution for crowdsourced data collection, algorithm training, and model validation suitable for resource-sensitive research and development environments. 

\section{Conclusion and Future Work}
The platform uniquely combines simulation, gamification, and adaptability with a low-cost physical test platform. This combination supports semi-supervised, crowdsourced data collection, rapid algorithm development cycles, and inexpensive model validation. 

There are opportunities for future improvement. For example, adding an in-game checkerboard pattern would allow us to evaluate the impact of camera calibration  on model performance. We plan to include simulated LIDAR to improve the simulator's utility for collision avoidance, and to create a track-builder utility or procedural track generator. Incorporating multiplayer, simulated traffic, and/or unpredictable events (``moose crossing??) would help train more complex scenarios. 

Because the simulator is based on a multi-platform game engine, broader distribution and the creation of improved scoring mechanisms and game modes will provide incentive for players to contribute informative supervised data, supporting rapid behavior cloning for long-tail events.  These same robust scoring metrics would allow us to rank the highest-performing drivers? training data more heavily than lower-scoring drivers when training the neural network. Some of this can be visible to the user (a ``disqualification?? notice), while other score metrics may be invisible (a hidden collider object could disable image and CSV capture and deduct from the user?s score when the vehicle leaves a ``safe?? region). This work will require developing a network backend for data storage and retrieval from diverse devices. 

There may also be opportunities to integrate the simulator and physical vehicles into an IoT framework\cite{doi:10.3846/16484142.2015.1079237} or with AR/VR tools\cite{ve.20181381}, such that physical vehicles inform the simulation in realtime and vise-versa.

Finally, crowdsourcing only works if tools are widely available. We hope to release the simulator and test platform details to the public once the models for both have been better-matched. 

\section{Acknowledgements}
We gratefully acknowledge the support of NVIDIA Corporation with the donation of the Titan Xp GPU used for this research.

\bibliographystyle{IEEEtran}
\bibliography{references}

\end{document}